%% file: example.tex
\documentclass[submission,copyright,creativecommons]{eptcs}

\input{preamble.tex}

\providecommand{\event}{ACT 2019} 
\usepackage{breakurl}             

\title{Learning Functors using Gradient Descent}
\author{Bruno Gavranović\footnote{Work done while the author was at TakeLab, FER, University of Zagreb, Croatia}
\institute{Mathematically Structured Programming Group \\
  University of Strathclyde \\
  Glasgow, UK}
\email{bruno@brunogavranovic.com}
}

\def\titlerunning{Learning Functors using Gradient Descent}
\def\authorrunning{Bruno Gavranović}
\begin{document}
\maketitle

\begin{abstract}
Neural networks are a general framework for differentiable optimization which includes many other machine learning approaches as special cases.
In this paper we build a category-theoretic formalism around a neural
network system called CycleGAN \cite{CycleGAN}.
CycleGAN is a general approach to unpaired image-to-image translation that has
been getting attention in the recent years.
Inspired by categorical database systems, we show that
CycleGAN is a ``schema'', i.e. a specific category presented by generators and
relations, whose specific parameter instantiations are just set-valued
functors on this schema. We show that enforcing \emph{cycle-consistencies}
amounts to enforcing composition invariants in this category.
We generalize the learning procedure to arbitrary such categories and show a
special class of \textit{functors}, rather than functions, can be learned using
gradient descent.
Using this framework we design a novel neural network system capable of learning to insert and delete objects from images without paired data.
We qualitatively evaluate the system on the CelebA dataset and obtain promising results.
\end{abstract}

\section{Introduction}

Compositionality describes and quantifies how complex things can be
assembled out of simpler parts. It is a principle which tells us that the design of abstractions in a system needs to be done in such a way that we can intentionally forget their
internal structure \cite{OnCompositionality}.
In the rapidly developing field of \emph{deep learning}, there are two interesting properties of neural networks related to compositionality: (i) they \textit{are} compositional -- increasing the number of layers tends to yield better performance, and (ii) they are discovering (compositional) structures in data.

\begin{figure}[H]
\begin{center}
   \includegraphics[width=0.5\textwidth]{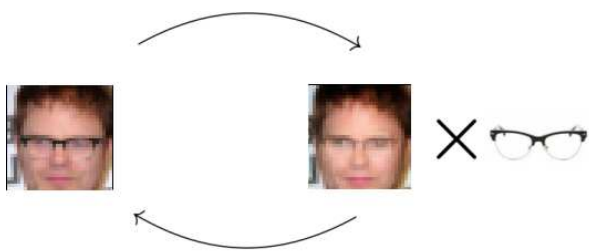}
   \caption{
     We devise a procedure to regularize neural network training when these networks are morphisms in a category presented by generators and relations.
   We use this regularization in the specific task of training neural networks to \textbf{remove glasses} from the
   face of a person and \textbf{insert them} parametrically. This is done
   without direct supervision and is object-invariant: the network is never
   provided with information that the image contains glasses or even that it
   contains a person.
 }
\label{fig:good_samples_montage}
\end{center}
\end{figure}

A modern deep learning system is made up of different types of components:
a neural network itself (a differentiable parameterized function; said to be
\emph{learning} during the process of optimization), an update rule, a cost/loss
function (a differentiable function used to direct the learning and assess the performance of the network), and also often overlooked data cleaning and
processing pipelines which supply the network with data.

In most deep learning setups the only component which is being optimized is,
unsurprisingly, the neural network itself. However, this has seen a change in
the recent years, where an increasing number of components of a modern deep
learning system started being \emph{modified during learning}. In many of these
cases, these other components have been replaced by other neural networks which
are trained and learned in parallel.
For instance, Generative Adversarial Networks \cite{GAN} also learn the \textit{cost
  function}. The paper \textit{Learning to Learn by gradient descent by gradient
descent} \cite{LTL} specifies networks that learn the \textit{update rule}. The
paper \textit{Decoupled Neural Interfaces using Synthetic Gradients}
\cite{SyntheticGradients} specifies how, surprisingly, even gradients
themselves can be learned: in this case networks are reminiscent of agents which
communicate information and gradient updates to each other.

These are just a few examples, but they give a sense of things to come.
As more and more components of these systems stop being fixed throughout
training, there is an increasingly larger need for more precise formal
specification of the things that \textit{do} stay fixed.
This is not an easy task; the invariants across all these networks seem to be
rather abstract and hard to describe.
In this paper we explore the hypothesis that the language of category theory could
be well suited to describe these systems in a compositional and precise manner.

\paragraph{Inter-domain mappings.}
Recent advances in neural networks describe the process of discovering high-level, abstract structure in data using
gradient information. As such, learning inter-domain mappings has received
increasing attention in recent years, especially in the context of
\textit{unpaired data} and image-to-image translation \cite{CycleGAN, AugmentedCycleGAN}.
\textit{Pairedness} of datasets $X$ and $Y$ generally refers to the existence of some
invertible function $X \rightarrow Y$. Building datasets that contain
that pairing information often requires extra human labor or computing resources.
Moreover, many aspects of \emph{human learning} do not involve paired datasets. As
eloquently described in the introduction of \cite{CycleGAN}, often we can reason
about stylistic differences between paintings of different painters, even though
never having seen paired data, i.e. \emph{the same} scene painted by two
different painters. This enables us to learn and generate high-level
information from data well beyond the capabilities of many of the current learning algorithms.

Motivated by the success of Generative Adversarial Networks (GANs) \cite{GAN} in image
generation, some existing unsupervised learning methods \cite{CycleGAN, AugmentedCycleGAN} use adversarial losses to learn the true data distribution of given domains of natural
images and \textit{cycle-consistency} losses to learn \textit{coherent} mappings between those domains.
CycleGAN is one of them. It is a system of neural networks which learns a one-to-one mapping between two domains.
Each domain has an associated \textit{discriminator}, while the mappings between
these domains correspond to \textit{generators}.
It includes two notions of learning: i) adversarial learning, where generators and
discriminators play the usual GAN minimax game \cite{GAN}, and ii) the
\emph{cycle-consistency learning}, where specific generator composition
invariants are enforced.
A commonly used example of this one-to-one mapping used in \cite{CycleGAN} is of
images of \textit{horses} and \textit{zebras}. Simply by changing the texture of
the animal in such an image we can, approximately, map back and forth between
these images. Learning \emph{how to change} this texture (without being provided
the information that there are horses or zebras in the image) is what CycleGAN
enables us to do.

\paragraph{Outline of the main contributions.}

In this paper we make the first steps of formalization of general systems based
on CycleGAN in the language of category theory.

We package CycleGAN into a category presented by generators and relations.
Given such a category -- which we call a \textit{schema}, inspired by
\cite{FunctorialDataMigration} -- we specify the architectures of its
constituent networks as a functor $\arch$.
We reason about various other notions found in deep learning, such as datasets,
embeddings, and parameter spaces.

We associate the training process with an indexed family of functors $\{ H_{p_i}: \fr
\rightarrow \set \}_{i=1}^T$, where $T$ is the number of training steps and $p$
is some choice of a parameter for that architecture.
Analogous to standard neural networks -- starting with a randomly
initialized $H_p$ we iteratively update it using gradient descent.
The optimization is guided by generalized version of \textit{two} objectives
found in \cite{CycleGAN}: adversarial minimax objective and the
cycle-consistency objective (also called here the \textit{path-equivalence objective}).

This approach yields useful insights and a large degree of generality: (i) it enables learning with unpaired data as it does not impose any constraints on ordering or pairing of
the sets in a category, and (ii) although specialized to generative models in
the domain of computer vision, this approach is domain-independent and general
enough to hold in any domain of interest, such as sound, text, or video.
Roughly, this allows us to think of a subcategory of $\set^{\fr}$ as a space in which we
can employ a gradient-based search.
In other words, we use specific network composition invariants as regularization during training, such that the imposed relationships guide the learning process.

We show that for specific choices of $\qfr$ and the dataset we recover GAN \cite{GAN} and CycleGAN \cite{CycleGAN}.
Furthermore, we describe a novel neural network system capable of learning to
remove and insert objects into an image with unpaired data (Figure \ref{fig:good_samples_montage}). We qualitatively evaluate the system on the CelebA dataset and obtain promising results.

\section{Towards Categorical Deep Learning}

Modern deep learning optimization algorithms can be framed as a gradient-based search in some function space $Y^X$, where $X$ and $Y$ are sets that have been endowed with extra structure.
Given some sets of data points $D_X \subseteq X$, $D_Y \subseteq Y$, a typical
approach for adding inductive bias relies on exploiting this extra structure.
This structure might be any sort of domain-specific features that can be exploited by various methods -- convolutions for images, Fourier transform for audio, and
specialized word embeddings for textual data.

In this paper we focus on a different sort of inductive bias - defined in
\cite{CycleGAN} - where the inductive bias is increased not by exploiting extra
structure of these sets, but rather by enforcing composition invariants of maps
between those sets.
We proceed by defining these schemas which contain the information about the
ways these maps can be composed.

\subsection{Model schema}\label{sec:model_schema}

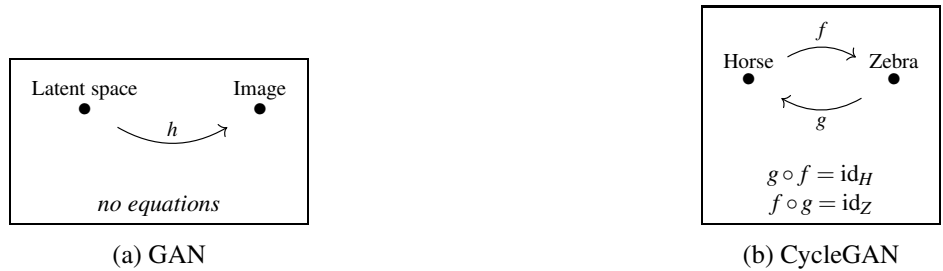
\begin{figure}[H]
  \begin{subfigure}[b]{0.5\columnwidth}
\centering
\boxCD{
\begin{tikzcd}[column sep=small, ampersand replacement=\&]
  \LTO{Latent space}\ar[rr, bend right, "h"]\&\&
  \LTO{Image} \\
\end{tikzcd}
  \\~\\\footnotesize
  \textit{no equations}
}
\caption{GAN}
  \end{subfigure}
  \hfill 
  \begin{subfigure}[b]{0.4\columnwidth}
\centering
\boxCD{
\begin{tikzcd}[column sep=small, ampersand replacement=\&]
  \LTO{Horse}\ar[rr, bend left, "f"]\&\&
    \LTO{Zebra}\ar[ll, bend left, "g"]\\
\end{tikzcd}
\\~\\\footnotesize
 $g \circ f  = \textrm{id}_H $ \\
 $f \circ g = \textrm{id}_Z  $
}
\caption{CycleGAN} 
  \end{subfigure}
\caption{Bird's-eye view of two popular neural network systems}
\label{fig:birdseye}
\end{figure}

Many deep learning models are complex systems, some comprised of several
neural networks. Each neural network can be identified with domain $X$, codomain $Y$, and
a \textit{differentiable parameterized function} $X \rightarrow Y$.
Given a \textit{collection} of such networks, we use a directed multigraph to
capture their interconnections.
Each directed multigraph $G$ gives rise to a corresponding free category on
that graph $\fr$. Based on this construction, Figure \ref{fig:birdseye} shows
the interconnection pattern for generators of two popular neural network
architectures: GAN \cite{GAN} and CycleGAN \cite{CycleGAN}.

Observe that CycleGAN has some additional properties imposed on it, specified by
equations in Figure \ref{fig:birdseye} (b). These are called cycle-consistency
conditions and can roughly be stated as follows: given domains $A$ and $B$
considered as sets, ${a \approx g(f(a)), \forall a \in A}$ and ${b \approx f(g(b)),
\forall b \in B}$. A particularly clear diagram of the cycle-consistency
condition can be found in \cite[Figure 3.]{CycleGAN}.

Our approach involves \textit{eta-reduction} of the aforementioned equations to
obtain ${id_a = g \circ f}$ and $id_b = f \circ g$. This allows us to package the newly formed equations as equivalence relations on the sets $\fr(A, A)$ and $\fr(B, B)$, respectively. This notion can be further packaged into a quotient category $\qfr$, together with the quotient functor ${\fr \xrightarrow{Q} \qfr}$.

This formulation of CycleGAN -- as a free category on a graph $G$ quotiented out
by a specific equivalence relation -- represents the
cornerstone of our approach. These schemas allow us to precisely reason only about the
interconnections between various concepts, while keeping any specific functions,
networks or other some other sets separate. All the other constructs in this
paper are structure-preserving maps between categories whose domain, roughly,
can be traced back to $\fr$.

\subsection{What is a neural network?}

In computer science, the idea of a \textit{neural network} colloquially means a
number of different things. At a most fundamental level, it can be interpreted as
a system of interconnected units called neurons, each of which has a firing
threshold acting as an information filtering system. Drawing inspiration
from biology, this perspective is thoroughly explored in literature. In many
other contexts we want to focus on the mathematical properties of a neural network and as such identify it with a function between sets $A \xrightarrow{f}
B$. Those sets are always equipped with some notion of smoothness (most commonly
Euclidean spaces). Functions are then considered to be maps of a given
differentiability class which preserve such structure.
We also frequently reason about a neural network jointly with its parameter space $P$
as a function of type $f: P \times A \rightarrow B$.

Without any loss of generality we illustrate how the learning in a neural
network is done with a simple example.
For instance, consider a classifier
in the context of supervised learning. An example is a convolutional neural network whose
input is a $32 \times 32$ \texttt{RGB} image and output is real number,
representing the probability of a cat appearing in the image. This network can be
represented as a function with the following type: $\RR^n \times \RR^{32 \times
  32 \times 3} \rightarrow \RR$, for some $n \in \mathbb{N}$. In this case
$\RR^n$ represents the parameter space of this network.

In the machine learning community, a function of such type is commonly referred
to as \emph{the neural network architecture}. It specifies an entire \textit{parameterized family} of functions of
type $\RR^{32 \times 32 \times 3} \rightarrow \RR$, because partial application
of each $p \in \RR^n$ yields a function $f(p, -) : \RR^{32 \times 32 \times 3}
\rightarrow \RR$.
This choice of a parameterized family of functions is part of the
\textit{inductive bias} we are building into the training process.
For example, in computer vision it is common to restrict the class of functions
to those that can be modeled by convolutional neural networks, while in natural
language processing it is common to restrict to those functions modeled by
recurrent neural networks.
Each of these functions can be evaluated on how much it agrees with the data points.
The process of learning, then, involves a priori specification of some such
function $f : \RR^n \times \RR^{32 \times 32 \times 3} \rightarrow \RR$ and some
initial $p_0 : \RR^n$. By measuring how much the function agrees with our data
points, we are able to ``wiggle'' that parameter $p_0$ and change it to the one that
gives us a function which slightly better agrees with the data points.

With this in mind, we recall the model schema.
For each morphism $A \rightarrow B$ in $\fr$ we are interested
in specifying a parameterized function $f : P \times A \rightarrow B$, i.e.~a parameterized
\textit{family of functions} in $\set$.
The function $f$ describes a neural network architecture, and a choice of a
partially applied $p \in P$ to $f$ describes a choice of some parameter value
for that specific architecture.

We capture the notion of parametrization with the category $\para$
\cite{BackpropAsFunctor}.
It is a strict symmetric monoidal category  whose objects are Euclidean spaces
and morphisms $\RR^n \rightarrow \RR^m$ are equivalence classes of differentiable functions of type $\RR^p \times \RR^n \rightarrow \RR^m$, for some $p$.
Composition of morphisms in $\para$ is defined in such a way that it explicitly
keeps track of parameters. For more details, we refer the reader to
\cite{BackpropAsFunctor}.

A closely related construction we use is $\euc$, the strict symmetric monoidal category
whose objects are finite-dimensional Euclidean spaces and morphisms are
differentiable maps. A monoidal product on $\euc$ is given by the cartesian
product.
We package both of these notions -- choosing an architecture and choosing
parameters -- into functors whose domain is $\fr$ and codomains are $\para$ and
$\euc$, respectively.

\subsection{Network architecture}

In the rest of the paper assume some directed multigraph $G$ has been specified and proceed to define some terminology.

\begin{definition}
We call a \emph{(neural network) architecture} any functor $\fr \rightarrow \para$ .
\end{definition}

To specify such a functor it is necessary to specify its action on objects in
$\fr$ and only on the generators of $\fr$ (since there are no relations).
Just as a choice of a single differentiable parameterized function is part of
the inductive bias we are building in to the network, so it follows that $\arch :
\fr \to \para$ (which consists of a family of such functions) is a choice of the inductive bias as well. For instance, one morphism in $\fr$ might get
mapped to one neural network, another morphism to another neural network. Their composition in $\fr$ is then mapped to the composite network.
For instance, both GAN and CycleGAN (Figure \ref{fig:birdseye}) will have their morphisms
mapped to specific convolutional networks.

Every choice of an architecture $\fr \xrightarrow{\arch} \para$ goes hand in hand
with the choice of a \textit{task embedding}.

\begin{proposition}
A \emph{task embedding} is a functor $\discr{\fr} \xrightarrow{E} \set$ defined as the composite 
$$
\vfr \to \fr \xrightarrow{\arch} \para \xrightarrow{U} \set
$$
where $U : \para \to \set$ is the forgetful functor mapping an Euclidean space
to the underlying set and a smooth map to its underlying function.
\end{proposition}

Task embedding is a useful notion in machine learning when we need to talk about
the space(s) in which our dataset(s) reside in. In most cases, we start out with
some dataset(s) already embedded in specific space(s) and thus our choice of
network architecture is limited - it needs to match the embedding at hand.

\subsection{Parameter space}\label{subsec:param_spec}

Each network architecture $f : \RR^n \times \RR^a \rightarrow \RR^b$ comes equipped
with its parameter space $\RR^n$.
Just as $\fr \xrightarrow{\arch} \para$ is a categorical generalization of
architecture, we now show there exists a categorical generalization of a
parameter space. In this case -- it is the parameter space of the functor $\fr \xrightarrow{\arch} \para$.
Before we move on to the main definition, we package the notion of parameter
space of a function $f : \RR^n \times \RR^a \rightarrow \RR^b$ into a simple function
$\mathfrak{p}(f) = \RR^n$.

\begin{definition}[Functor parameter space]\label{def:parameter_space}
Let $\gen{\fr}$ the set of generators in $\fr$. The total parameter map
$\mathcal{P}: Ob(\para^{\fr}) \rightarrow \euc$ is a function that assigns to each functor $\fr \xrightarrow{\arch} \para$
the product of the parameter spaces of all its generating morphisms:
\[
\mathcal{P}(\arch) := \displaystyle \prod_{f \in \gen{\fr}}{\mathfrak{p}(\arch(f))}
\]
\end{definition}
Essentially, just as $\mathfrak{p}$ returns the parameter space of a function,
$\mathcal{P}$ does the same for a \textit{functor}.

We are now in a position to talk about parameter specification. Recall the
non-categorical setting: given some network architecture ${f: P \times A \rightarrow B}$ and a choice of $p \in \mathfrak{p}(f)$ we can partially apply
the parameter $p$ to the network to get $f(p, -) : A \rightarrow B$.
This admits a straightforward generalization to the categorical setting.

\begin{definition}[Parameter specification]\label{def:param_spec}
Parameter specification $\pspec$ is a dependently typed function with the
following signature:
\begin{equation}
  (\arch : Ob(\para^{\fr})) \times \mathcal{P}(\arch) \rightarrow Ob(\euc^{\fr})
\label{eq:param_spec}
\end{equation}
Given an architecture $\arch$ and a parameter choice $(p_f)_{f \in \gen{\fr}} \in \mathcal{P}(\arch)$ for that architecture, it defines a choice of
a functor in $\euc^{\fr}$. This functor acts on objects the same as $\arch$. On
morphisms, it partially applies every $p_f$ to the corresponding morphism
$\arch(f) : \RR^n \times \RR^a \rightarrow \RR^b$, thus yielding $f(p_f, -) : \RR^a \rightarrow \RR^b$ in $\euc$.
\end{definition}

Elements of $\euc^{\fr}$ will play a central role later on in the paper.
These elements are functors which we will call \textit{Models}. Given some
architecture $\arch$ and a parameter $p \in \ps(\arch)$, a model $\fr \xrightarrow{\model_p} \euc$
generalizes the standard notion of a model in machine learning -- it can be used
for inference, evaluated and it has parameters which can be updated during training. 

Analogous to database instances in \cite{FunctorialDataMigration}, for a given
schema $\fr$, we call a \textit{network instance} the functor $H_p : \fr \to
\set$ defined as the composite
$$
\fr \xrightarrow{Model_p} \euc \xrightarrow{U} \set
$$

\subsection{Path equivalence relations}\label{subsec:path_equiv}

So far, we have been only considering schemas given by $\fr$. This indeed is a limiting
factor, as it assumes the categories of interest are only those without any
imposed relations. One example of a schema
\textit{with} relations is the CycleGAN schema (Figure \ref{fig:birdseye} (b)) where we
are enforcing some composition invariants initially not present in $\fr$.

This is done via some quotient functor $Q: \fr \to \qfr$, where $\qfr$ is the
quotient category whose objects are objects of $\fr$ and morphisms are
equivalence classes of morphisms in $\fr$.
However, it has to be noted that some arbitrary parameter instantiations on
$\qfr$ will not yield a functor $\qfr \to \set$ (i.e. a network instance) since
we cannot guarantee this functor will preserve composition.

However, working \emph{only} with functors $\fr \to \set$ and iteratively
updating them where updates penalize discrepancies between the imposed
relations, we will show there are cases where it is possible for the training
process to converge to a functor $H : \fr \to \set$ which actually preserves
these relations.
There is a general statement about quotient categories
(\cite{WorkingMathematician}, Section 2.8., Proposition 1.) which tells us that in such a case, $H$ induces a unique $H' : \qfr \to \set$ such that the following diagram commutes:

\begin{figure}[H]
\centering
\begin{tikzcd}[column sep = 30pt, row sep = 30pt]
\free(G) \arrow[rd, "\Fun{H}"] \arrow[d, "\Fun{Q}"'] &  \\
\qfr \arrow[r, dashed, "\Fun{H'}"']
& \Cat{Set}
\end{tikzcd}
\caption{Functor $H$ which preserves path-equivalence relations factors uniquely
  through $Q$. }
\label{fig:quot_cat}
\end{figure}

In other words, this allows us to initially guess a map $\qfr \to \set$ which is
\emph{not a functor} and incentivize the learning algorithm to \emph{learn a
  functor} using gradient descent. This describes how learning schemas with arbitrary relations fits into the categorical framework.

\section{Data}

We have described constructions which allow us to pick an architecture for a
schema and consider its different models $\model_p$, each of them identified with a
choice of a parameter $p \in \ps(\arch)$.
In order to understand how the optimization process is steered in
updating the parameter choice for an architecture, we need to understand
a vital component of any deep learning system -- datasets themselves. 

This necessitates that we also understand the relationship between datasets
and the space they are embedded in.

\begin{definition}\label{def:datasetf}
  Let $\discr{\fr} \xrightarrow{E} \set$ be some embedding.
  We call a \textbf{dataset} any subfunctor of $E$.
\end{definition}

In other words, some subfunctor $\Fun{D}_E: \vfr \rightarrow \set$ of $E$ has
the semantics of dataset because it maps each object $A \in Ob(\fr)$ to a
dataset $\Fun{D}_E(A) := \{a_i\}_{i=1}^N \subseteq \Fun{E}(A)$ of some kind
which is embedded in $E(A)$.

Note that we refer to this functor in the singular, although it assigns a
dataset to \textit{each} object in $\fr$. We also highlight that the domain of
$\Fun{D}_E$ is $\vfr$, rather than $\fr$. We generally cannot provide an action on
morphisms because datasets might be incomplete.
Going back to the example with Horses and Zebras -- a dataset functor on $\fr$ in
Figure \ref{fig:birdseye} (b) maps $\mathrm{Horse}$ to the set of obtained horse images
and $\mathrm{Zebra}$ to the set of obtained zebra images.

The subobject relation $D_E \subseteq E$ in Proposition \ref{def:datasetf} reflects an important property of data;
we cannot obtain some data without it being in some shape or form, embedded in
some larger space. Any obtained data thus implicitly fixes an embedding.

Observe that when we have a dataset in standard machine learning, we have a
dataset \textit{of something}. We can have a dataset of historical weather data,
a dataset of housing prices in New York or a dataset of cat images.
What ties all these concepts together is that each element $a_i$ of some
dataset $\{a_i\}_{i=1}^N$ is an instance of a more general concept. As a trivial
example, every image in the dataset of horse images is a
\textit{horse}. The word \textit{horse} refers to a more general concept and as
such could be generalized from some of its instances which we \textit{do not
  possess}. But all the horse images we possess are indeed an example of a horse. By considering everything to be embedded in some space $E(A)$ we capture this statement with the relation $\{a_i\}_{i=1}^N \subseteq \cpt(A) \subseteq E(A)$. Here $\cpt(A)$ is the set of all instances of some notion $A$ which are embedded in $E(A)$. In the running example this corresponds to all images of horses in a given space, such as the space of all $64 \times 64$ \texttt{RGB} images. Obviously, the precise specification of $\cpt(A)$ is unknown -- as we cannot enumerate or specify the set of \textit{all} horse images.

We use such calligraphy to denote this is an abstract concept.
Despite the fact that its precise specification is unknown, we can still reason about its relationship to other structures.
Furthermore, as it is the case with any abstract notion, there might be some
edge cases or it might turn out that this concept is ambiguously defined or
even inconsistent. Moreover, it might be possible to identify a dataset with
multiple concepts; is a dataset of male human faces associated with the concept of
male faces or is it a non-representative sample of all faces in general?
We ignore these concerns and assume each dataset is a dataset of some
well-defined, consistent and unambiguous concept. This does not change the
validity of the rest of the formalism in any way as there exist plenty of
datasets satisfying such a constraint.

Armed with intuition, we show this admits a generalization to the
categorical setting.
Just as $\{a_i\}_{i=1}^N \subseteq \cpt(A) \subseteq
E(A)$ are all subsets of $E(A)$ we might hypothesize the domain of $\cpt$ is
$\vfr$ and that $D_E \subseteq \cpt \subseteq E$ are all subfunctors of $E$. However, just as we assign a set of all concept instances to \textit{objects} in $\fr$, we also assign a function between these sets to \textit{morphisms} in $\fr$. Unlike with datasets, this can be done because, by definition, these sets are not incomplete.

\begin{definition}
  Given a schema $\qfr$ and a dataset $\vfr \xrightarrow{D_E} \set$, a
  \textbf{concept} associated with the dataset $D_E$ embedded in $E$ is a
  functor $\cpt: \qfr \rightarrow \set$ such that $D_E \subseteq \cpt \circ I
  \subseteq E$. We say $\cpt$ picks out sets of concept instances and functions
  between those sets. 
\end{definition}

Another way to understand a concept $\qfr \xrightarrow{\cpt} \set$ is that it is
required that a human observer can tell, for each $A \in Ob(\fr)$
and some $a \in \emb(A)$ whether $a \in \cpt(A)$. Similarly for morphisms, a
human observer should be able to tell if some function $\cpt(A) \xrightarrow{f}
\cpt(B)$ is an image of some morphism in $\qfr$ under $\cpt$.

\begin{example}
Consider the GAN schema in Figure \ref{fig:birdseye} (a) where
$\cpt(\mathrm{Image})$ is a set of all images of human faces embedded in some
space such as $\RR^{64 \times 64 \times 3}$. For each image in this space, a human observer should be able to tell if that image contains a face or not.
We cannot enumerate such a set $\cpt(\mathrm{Image})$ or write it down
explicitly, but we can easily tell if an image contains a given concept.
Likewise, for a morphism in the CycleGAN schema (Figure \ref{fig:birdseye} (b)),
we cannot explicitly write down a function which transforms a horse into a
zebra, but we can tell if some function did a good job or not by testing it on
different inputs.
\end{example}

The most important thing related to this concept is that this represents the
goal of our optimization process. Given a dataset $\vfr \xrightarrow{D_E} \set$,
want to extend it into a functor $\qfr \xrightarrow{\cpt} \set$, and actually
\textit{learn} its implementation.

\subsection{Restriction of network instance to the dataset}

We have seen how data is related to its embedding. We now describe the
relationship between \textit{network instances} and data.

Observe that network instance $H_p$ maps each object $A \in Ob(\fr)$ to the
entire embedding $H_{p_i}(A) = E(A)$, rather than just the concept $\cpt(A)$.
Even though we started out with an embedding $E(A)$, in generative data
modelling we are usually interested in restriction of that set just to the set of instances corresponding to some concept $A$.

For example, consider a diagram such as the one in Figure \ref{fig:birdseye}
(a). Suppose the result of a successful training was a functor $\fr
\xrightarrow{H} \set$.  Suppose that the image of $h : \LTO{Latent space} \to \LTO{Image}$ is $H(h) : \zo \rightarrow \gimg$. As such, our interest is mainly the
restriction of $\gimg$ to $\cpt(\mathrm{Image})$, the image of $\zo$ under $H(h)$, rather than the
entire $\gimg$. In the case of horses and zebras in Figure \ref{fig:birdseye} (b), we are interested in a map
$\cpt(\mathrm{Horse}) \rightarrow \cpt(\mathrm{Zebra})$ rather than a map $\gimg
\rightarrow \gimg$.
In what follows we show a construction which restricts some $H_p$ to its
smallest subfunctor which contains the dataset $D_E$.
Recall the previously defined inclusion 
\begin{definition}
Let $\Fun{D}_E: \vfr \rightarrow \set$ be a \textbf{dataset}. Let $\fr
\xrightarrow{H_p} \set$ be a network instance on $\fr$. The \textbf{restriction}
of $H_p$ to $D_E$ is a subfunctor of $H_p$ defined as follows:
\[
\ih := \bigcap\limits_{\{G \in
    Sub(H_p)) \mid \Fun{D}_E \subseteq \Fun{G} \circ \Fun{I}\}}G
\]
where $Sub(H_p)$ is the set of subfunctors of $H_p$, and $\vfr
\xhookrightarrow{I} \fr$ is the inclusion.
\end{definition}

This definition is quite condensed so we supply some intuition.
We first note that the meet is well-defined because each $G$ is a subfunctor of $H$.
In Figure \ref{fig:data_functors} we depict the newly defined constructions
using a commutative diagram.

\tikzset{
    labl2/.style={anchor=south, rotate=-30, inner sep=.5mm}
}
\begin{figure}[H]
\centering
\begin{tikzcd}[column sep=30pt, row sep=40pt]
  & \fr \arrow[dr, "\mathbin{\rotatebox[origin=c]{40}{$\subseteq$}}", phantom,
  bend left=10] \arrow[dr, "H_p", bend left] \arrow[dr, "\ih", bend right=28] & \\
  \vfr \arrow[ru, hook, "I"]  \arrow[rr, "D_E"] & & \set
\end{tikzcd}
\caption{The functor $\ih$ is a subfunctor of  $H_p$ and $D_E$ is a subfunctor of $\ih \circ I$.}
\label{fig:data_functors}
\end{figure}
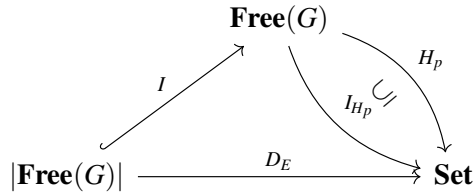

It is useful to think of $I_H$ as a restriction of $\Fun{H}$ to the
\textit{smallest} functor which fits all data and mappings between the data.
This means that $\ih$ contains all data samples specified by $D_E$. \footnote{We
also note vague similarities to the notion of a \emph{Kan Extension}.}

\section{Optimization}

We now describe how data guides the search process. We identify
the goal of this search with the concept functor $\qfr
\xrightarrow{\cpt} \set$. This means that given a schema $\qfr$ and data
$\vfr \xrightarrow{D_E} \set$ we want to train some architecture $\arch$ and
find a functor ${\qfr \xrightarrow{H'} \set}$ that can be identified with $\cpt$. Of
course, unlike in the case of the concept $\cpt$, the implementation of $H'$ is
something that will be known to us. We proceed by defining the notion of a \textit{task} which includes all the necessary information to employ a gradient-based search.

\begin{definition}\label{def:task}
A task is a 5 tuple $(G, \sim, E, D_E, \cpt)$, where $G$ is a directed
multigraph, $\sim$ a congruence relation on $\fr$ and the rest are functors: $\vfr \xrightarrow{E} \set$, $\vfr \xrightarrow{D_E} \set$, and ${\qfr \xrightarrow{\cpt} \set}$.
\end{definition}

Moreover, observe that an embedding $E$ too, as $D_E \subseteq E$ in turn also
narrows our choice of architecture $\fr \xrightarrow{\arch} \para$, which it has to agree with the embedding on objects. This situation fully
reflects what happens in standard machine learning practice -- a neural network
$P \times A \rightarrow B$ has to be defined in such a way that its domain $A$
and codomain $B$ embed the datasets of all of its inputs and outputs, respectively. Even though for the same schema $\qfr$ we might want to consider different datasets, we will always assume a chosen dataset corresponds to a single training goal $\cpt$.

\subsection{Optimization objectives}

We generalize the training procedure described in \cite{CycleGAN} in a natural
way, free of ad-hoc choices.

Suppose we have a task $(G, \sim, E, D_E, \cpt)$.
After choosing an architecture $\fr \xrightarrow{\arch}
\para$ consistent with the embedding $E$ and with the right inductive bias, we start with a randomly chosen parameter
$\theta_0 \in \ps(\arch)$. This amounts to a choice of a specific $\fr \xrightarrow{\model_{\theta_0}} \euc$.
Using the loss function defined further down in this section, we partially
differentiate each $f : \RR^n \times \RR^a \rightarrow \RR^b \in \gen{\fr}$ with
respect to the corresponding $p_f$ . We then obtain a new parameter value for
that function using some update rule, such as Adam \cite{Adam}.
The product of these parameters for each of the generators $(p_f)_{f \in
  \gen{\fr}}$ (Definition \ref{def:parameter_space}) defines a new parameter
$\theta_1 \in \ps(\arch)$ for the model $\model_{\theta_1}$.
This procedure allows us to iteratively update a given $\model_{\theta_i}$ and
as such fixes a sequence $\{\theta_0, \theta_1, \dots, \theta_T\}$ on some subset of $\ps(\arch)$.

Now we describe the optimization objective using a loss function. The loss
function will be a weighted sum of two components: the \textit{adversarial loss}
and the \textit{path-equivalence} loss. As we slowly transition to standard
machine learning lingo, we note that some of the notation here will be untyped
due to the lack of a deeper categorical understanding of these concepts.\footnote{
  Categorical formulation of the adversarial component of Generative Adversarial
  Networks is still an open problem. It seems to require
  nontrivial reformulations of existing constructions \cite{BackpropAsFunctor}
  and at least a partial integration of Open Games \cite{CompositionalGameTheory} into the framework of gradient-based optimization.}

We start by assigning a discriminator to each object $A \in Ob(\fr)$ using the following function:
\begin{equation*}
\mathbf{D} : (A : Ob(\fr)) \rightarrow \para(\arch(A), \RR)
\label{eq:discr_function}
\end{equation*}

This function assigns to each object $A \in Ob(\fr)$ a morphism in $\para$ such that its
domain is that given by $\arch(A)$. This will allow us to compose compatible
generators and discriminators.
For instance, consider $\arch(A) = \RR^a$. Discriminator $\bd(A)$ is then a
function of type $\RR^q \times \RR^a \rightarrow \RR$ and an element of  $\para(\RR^a, \RR)$,
where $\RR^q$ is the parameter space of the discriminator.
As a slight abuse of notation -- and to be more in line with machine learning
notation -- we will call $\bd_A$ discriminator of the object $A$ with some partially applied parameter value $\bd(A)(p, -)$.

In the context of GANs, when we refer to a generator we refer to the image of a generating morphism in $\fr$ under $\arch$.
Similarly, as with discriminators, a generator corresponding
to a morphism $\RR^a \xrightarrow{f} \RR^b$ in $\para$ with some partially applied
parameter value will be denoted using $\bg_f$.

The GAN minimax objective $\mathcal{L}_{GAN}^B$ for a generator $\bg_f$ and a
discriminator $\bd_B$ is stated in Eq. \eqref{eq:wgan-gp}. In
this formulation we use the Wasserstein distance \cite{WGAN}.
The generator is trained to minimize the loss in the 
Eq. \eqref{eq:wgan-gp}, while the discriminator is trained to maximize it.

\begin{equation}
  \begin{split}
\mathcal{L}_{GAN}^B(\bg_f, \bd_B) & := \mathop{\mathbb{E}}_{b \sim D_E(B)} 
\left[ \bd_B(b) \right] \\
    &- \mathop{\mathbb{E}}_{a \sim D_E(A)} \left[ \bd_B(\bg_f(a))  \right]
  \end{split}
\label{eq:wgan-gp}
\end{equation}

The second component of the total loss is a generalization of
\textit{cycle-consistency loss} in CycleGAN \cite{CycleGAN}, analogous to the
generalization of the cycle-consistency condition in Section \ref{sec:model_schema}.

\begin{definition}
 Let $A \doublerightarrow{f}{g} B$ be two morphisms in $\fr$ and suppose $f \sim
 g$. Let $\mathsf{Model}_i : \fr \to \euc$ be a model. Then there is a \textbf{path equivalence loss}
 $\mathcal{L}_{\sim}^{f, g}$   defined as:
\begin{equation*}
\mathcal{L}_{\sim}^{f, g} := \mathbb{E}_{a \sim
   D_E(A)} \big[ {\vert \vert \model_i(f)(a) - \model_i(g)(a) \vert \vert}_1
 \big]
\label{eq:path_equiv_loss}
\end{equation*}

\end{definition}

When this loss is zero, a unique functor $H' : \qfr \to \set$ will exist that makes the corresponding
diagram commute (as detailed in subsection \ref{subsec:path_equiv}). These two loses enable us to state the total loss simply as a weighted sum of adversarial losses for all generators and path equivalence losses for all equations.

\begin{definition}\label{def:total_loss}
The \textbf{total loss} is given as the sum of all adversarial and path
equivalence losses:
\begin{equation*}
\loss_i := \displaystyle\sum_{A \xrightarrow{f} B \in \gen{\fr}}{\loss_{GAN}^B(\bg_f, \bd_B)} + \gamma\sum_{f \sim g}{\mathcal{L}_{\sim}^{f, g}}
\label{eq:total_loss}
\end{equation*}

where $\gamma$ is a hyperparameter that balances between the adversarial loss
and the path equivalence loss.
\end{definition}


\subsection{Functor space}\label{sec:functor_space}

Given an architecture $\arch$, each choice of $p \in \ps(\arch)$ specifies a functor of type $\fr \rightarrow \set$.
In this way exploration of the parameter space amounts to exploration of part
of the functor category $\set^{\fr}$.
Roughly stated, this means that a choice of an architecture adjoins a notion of
\textit{space} to the image of $\pspec(\arch, -)$ in the functor category
$\set^{\fr}$. This space inherits all the properties of $\euc$.

By using gradient information to search the parameter space $\ps(\arch)$, we
are effectively using gradient information to search part of the functor space
$\set^{\fr}$. Although we cannot explicitly explore just $\set^{\qfr}$,
we penalize the search method for veering into the parts of this space where the
specified path equivalences do not hold. As such, the inductive bias of the
model is increased without special constraints on the datasets or the embedding space -
we merely require that the space is differentiable and that is has a sensible notion of
distance.

Note that we do not claim inductive bias is \textit{sufficient} to guarantee
training convergence, merely that it is a useful regularization method applicable to
a wide variety of situations.
As categories can encode complex relationships between concepts and as functors map
between categories in a structure-preserving way -- this enables
\textit{structured learning} of concepts and their interconnections in a very
general fashion.

\section{Product task}\label{sec:product_schema}

We now present a choice of a dataset for the CycleGAN schema which
makes up a novel task we will call \textit{the product task}. 
The interpretation of this task comes in two flavors: as a simple change of
dataset for the CycleGAN schema and as a method of composition and decomposition
of images.

Just as we can take the product of two real numbers $a, b \in \RR$ with a
multiplication function $(a, b) \mapsto ab$, we show we can take a product of
some two sets of images $A, B \in \set$ with a neural network of type $A \times B
\rightarrow AB$. We will show $AB \in \set$ is a set of images which possesses
all the properties of a categorical product. 

The categorical product $A \times B$ is uniquely isomorphic to any other object $AB$ which satisfies the universal property of the categorical product of objects $A$ and $B$.
This isomorphism will be central to the notion of the product task.
Recall that in a cartesian category such as $\set$ there already exists a notion of a
categorical product -- the cartesian product. 
Namely, we will show that there are cases where it is possible to (in addition
to $A \times B$) specify another object $AB$ which can be interpreted as a
categorical product isomorphic to $A \times B$.
When $A$ and $B$ are images containing some objects, $AB$ can be interpreted
as a semantic combination of two objects, perhaps in a non-trivial way.
Of course, this isomorphism only exists when
no information is lost combining two images, but we will see that, practically, even with
some loss of information, the results are still interesting and useful.

For instance, if $A$ are images of glasses and $B$ are images of people, then
$AB$ are images of people wearing glasses. For each image of a person $a \in A$
and glasses $b \in B$, there is an image $ab \in AB$ of person $a$ wearing
glasses $b$. 

Furthermore, $AB$ being a categorical product implies existence of the
projection maps ${\theta_A : AB \to A}$ and $\theta_B : AB \to B$.
This is where the difference from a cartesian product becomes more apparent.
The domain of the corresponding projections $\theta_A$ and $\theta_B$ is not a simple pair of objects $(a, b)$ and thus these projections cannot merely discard an element. $\theta_A$ needs to learn to remove $A$ from a potentially complex domain. As such, this can be any complex, highly non-linear function which satisfies coherence conditions of a categorical product.

We will be concerned with supplying this new notion of the product $AB$ with a
dataset and learning the image of the isomorphism $AB \cong A
\times B$. We
illustrate this on a concrete example. Consider a dataset $A$ of images of human
faces, a dataset $B$ of images of glasses, and a dataset $AB$ of people
\textit{wearing} glasses. Learning this isomorphism amounts to learning two
things: (i) learning how to decompose an image of a person wearing glasses $(ab)_i$ into an image of
a person $a_j$ and image $b_k$ of these glasses, and (ii) learning how to map this person $a_j$ and some other glasses $b_l$ into an image of a person $a_j$ wearing glasses $b_l$. Generally, $AB$ represents some sort of composition of objects $A$ and $B$ in the image space such that all information about $A$ and $B$ is preserved in $AB$. Of course, this might only be approximately true. Glasses usually cover a part of a face and sometimes its dark shades cover up the eyes -- thus losing information about the eye color in the image and rendering the isomorphism invalid. However, in this paper we ignore such issues and assume that the networks $\arch(d)$ can learn to unambiguously fill part of the face where the glasses were and that $\arch(c)$ can learn to generate and superimpose the glasses on the relevant part of the face. 

Even though for the product task we fix the same graph $G$ as in the CycleGAN
(and thus same CycleGAN schema from Figure \ref{fig:birdseye} (b)), we label one of its objects as $AB$ and the other one as
$A \times B$. Note that this does not change the schema itself, the labeling is
merely for our convenience. The notion of a product or its projections is not captured in the schema itself.
As schemas are merely categories presented with generators $G$ and relations
$R$, they lack the tools needed to encode a complex abstraction such as a
universal construction. \footnote{We note a high similarity with a notion of a
  \emph{sketch} \cite{QueryLiftingProblems}, but do not explore this connection further.} So how do we capture the notion of a product?

In this paper we frame this simply as a specific dataset
functor $\fr \to \set$, which we now describe. 
A dataset functor corresponding to the product task maps the object $A \times B$ in
CycleGAN schema to a \textit{cartesian product of two datasets}, $D_E(A \times B) =
\{a_i\}_{i=0}^N \times \{b_j\}_{j=0}^M$. It maps the object $AB$ to a dataset
$\{(ab)_i\}_{i=0}^N$. In this case $ab$, $a$, and $b$ are free to be any
elements of datasets of a well-defined concept $\cpt$.
Although the difference between the product task and the CycleGAN task boils down to a different choice of a dataset functor, we note this is a key aspect which allows for a significantly different interpretation of the task semantics.

By considering $A$ as some \textit{image background} and $B$ as the
\textit{object} which will be inserted, this allows us to interpret $d$ and $c$
as maps which \textit{remove an object from the image} and \textit{insert an
  object} in an image, respectively. 
This seems like a novel method of object generation and deletion with unpaired data, though we cannot claim to know the literature well enough to be sure.

\section{Experiments}\label{experiments}

In this section we test whether the product task described in Section
\ref{sec:product_schema} can be trained in practice. In our experiments we use
the CelebA dataset. CelebFaces Attributes Dataset (CelebA) \cite{CelebA} is a large-scale face attributes dataset with more than 200000 celebrity images and cover large pose variations and background clutter. Frequently used for image generation purposes, it fits perfectly into the proposed paradigm of the product task. Each image is equipped with 40 attribute annotations, which include ``eyeglasses'', ``bangs'', ``pointy nose'', ``wavy hair'' etc., as simple boolean flags.

We used these attribute annotations to separate CelebA into two datasets: the
dataset $D_E(AB)$ consisting of images with the attribute ``Eyeglasses'' and the
dataset $D_E(A)$ consisting of all the other images. Given that we could not obtain a dataset of images of \textit{just glasses}, we set $D_E(B_Z) = \zo$ and add the subscript $Z$ to $B$, as to make it more clear we are not generating images of this object. We refer to an element $z \in D_E(B_Z)$ as a \textit{latent vector}, in line with machine learning terminology. This is similar to usual generative modelling with GANs where the input is vector from some latent space. This is a parametrization of all the missing information from $A$ such that $A \times B_Z \cong AB$.

We investigated three things: (i) whether it is possible to
\textit{generate an image of a specific person wearing specific glasses}, (ii)
whether we can \textit{change} glasses that a person wears by changing the
corresponding latent vector, and (iii) whether the same latent vector
corresponds to the same glasses, irrespectively of the person we pair it with.
We leave the implementation details of training neural networks for these
experiments to the appendix and here only describe the results.
The only metric we use here to gauge the performance of these networks (other
than the value of the generator/discriminator losses) is visual
inspection of generated images.

\subsection{Results}

\begin{figure}
  \centering
\begin{subfigure}[t]{0.47\columnwidth}
        \centering
         {\includegraphics[width=0.95\columnwidth]{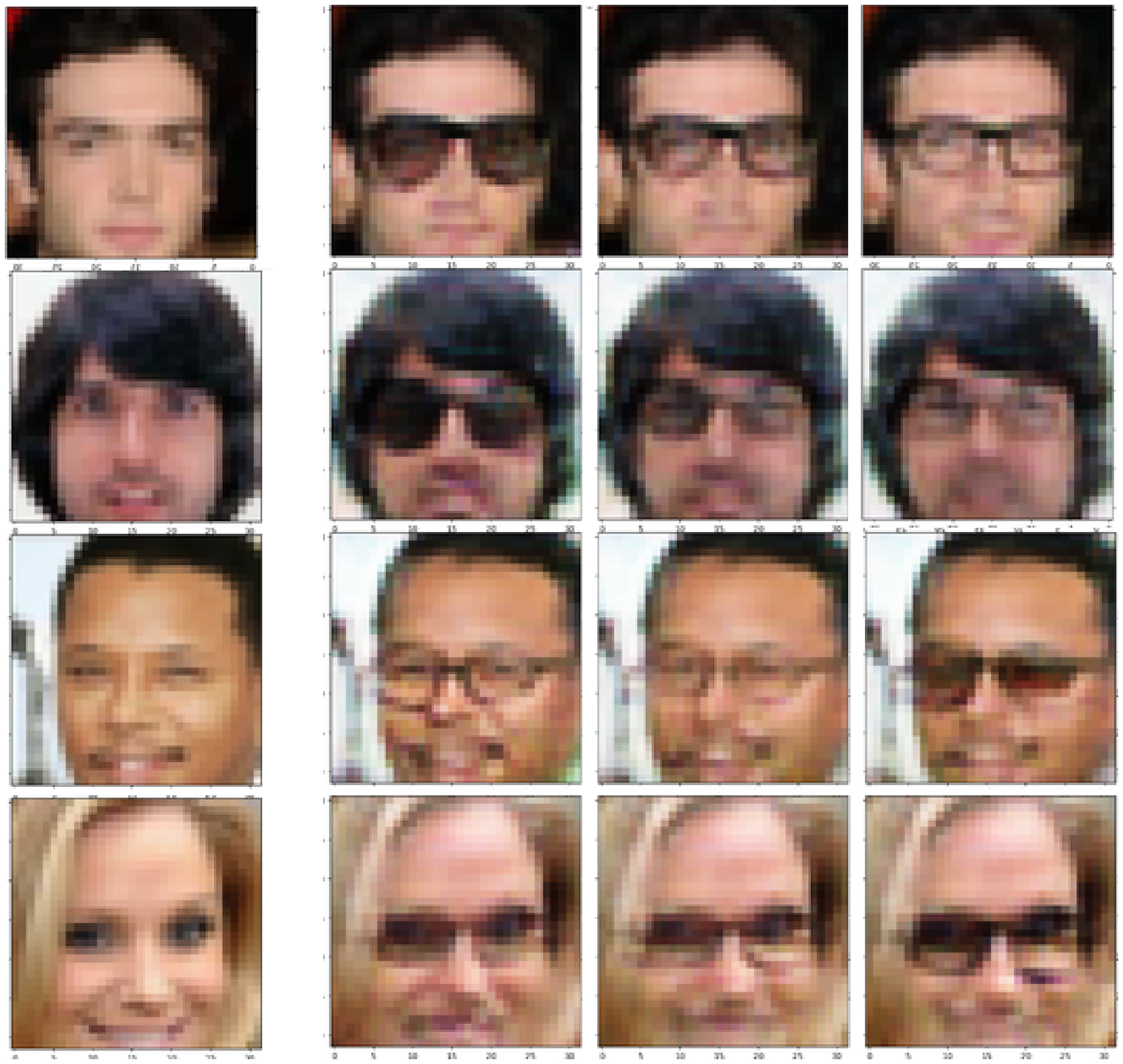}}
        \caption{}
    \end{subfigure}%
\begin{subfigure}[t]{0.47\columnwidth}
        \centering
         {\includegraphics[width=0.92\columnwidth]{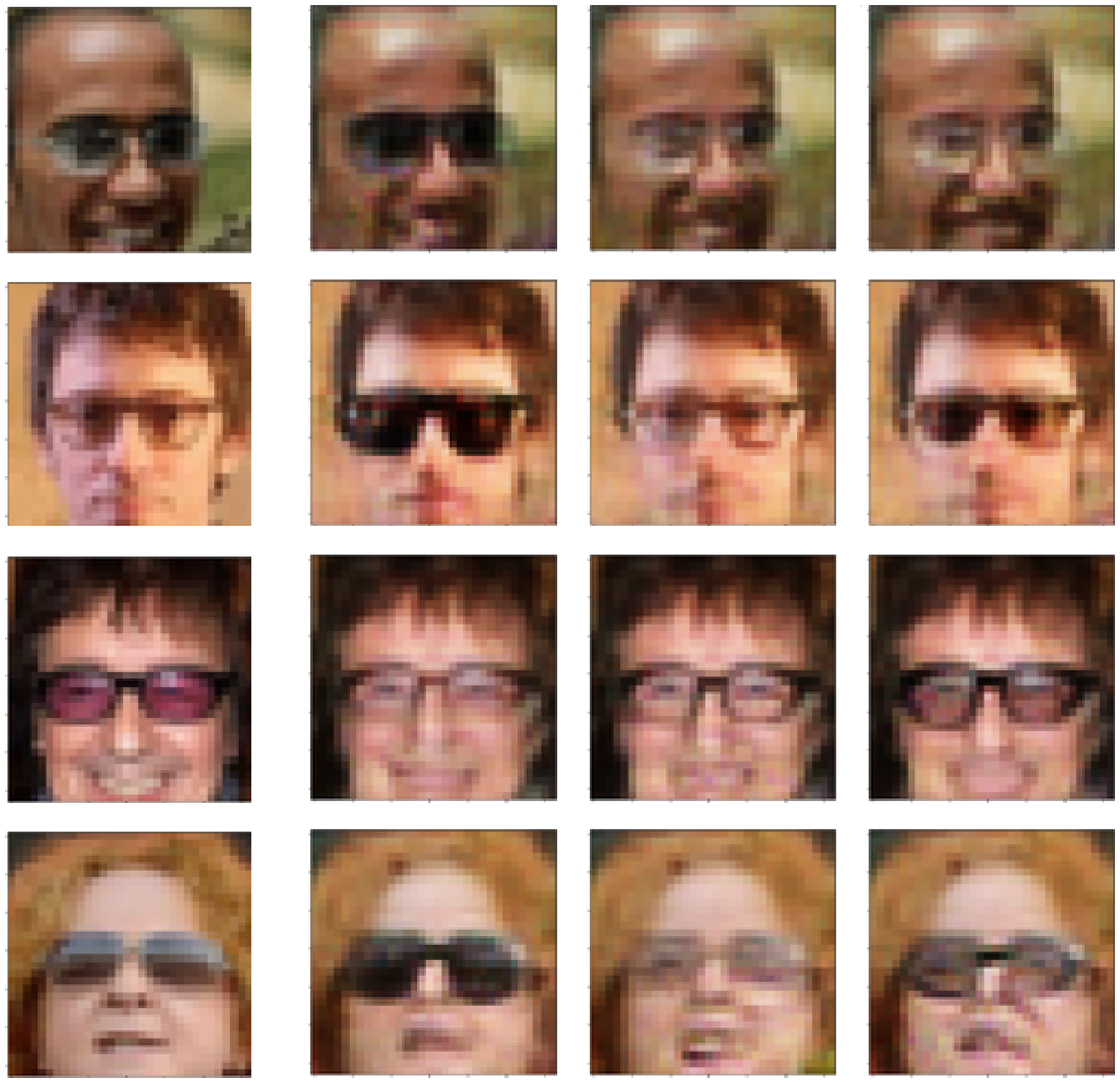}}
        \caption{}
    \end{subfigure}%
 \caption{Parametrically \textit{adding} glasses (a) and \textit{changing}
   glasses (b) on a person's face. (a): the leftmost column shows a sample
   from the dataset $a_i \in D_E(A)$. Three rightmost columns show the
   result of $c(a_i, z_j)$, where $z_j \in D_E(B_Z)$ is a randomly sampled latent
   vector. (b): leftmost column shows a sample from the dataset $(ab)_i \in
   D_E(AB)$. Three rightmost columns show the image $c(\pi_A(d((ab_i))), z_j)$
   which is the result of changing the glasses of a person. The latent vector
   $z_j \in D_E(B_Z)$ is randomly sampled.}
\label{fig:add_change_glasses}
\end{figure}

\begin{figure}[H]
\begin{minipage}[t]{0.45\linewidth}
      {\includegraphics[width=\columnwidth]{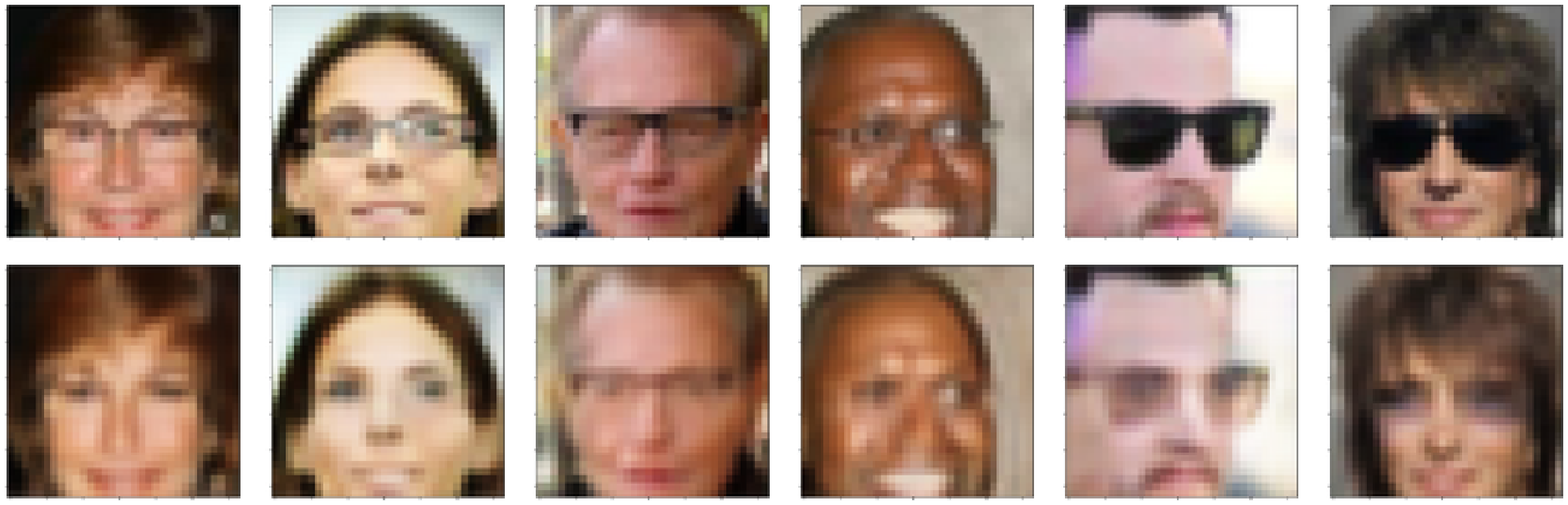}}
      \caption{Top row shows samples ${(ab)_i \in D_E(AB)}$. Bottom
        row shows the result of a function $\pi_A \circ d: AB \rightarrow A$
        which removes the glasses from the person.}
      \label{fig:remove_glasses}
\end{minipage}\qquad
\begin{minipage}[t]{0.45\linewidth}
      {\includegraphics[width=\columnwidth]{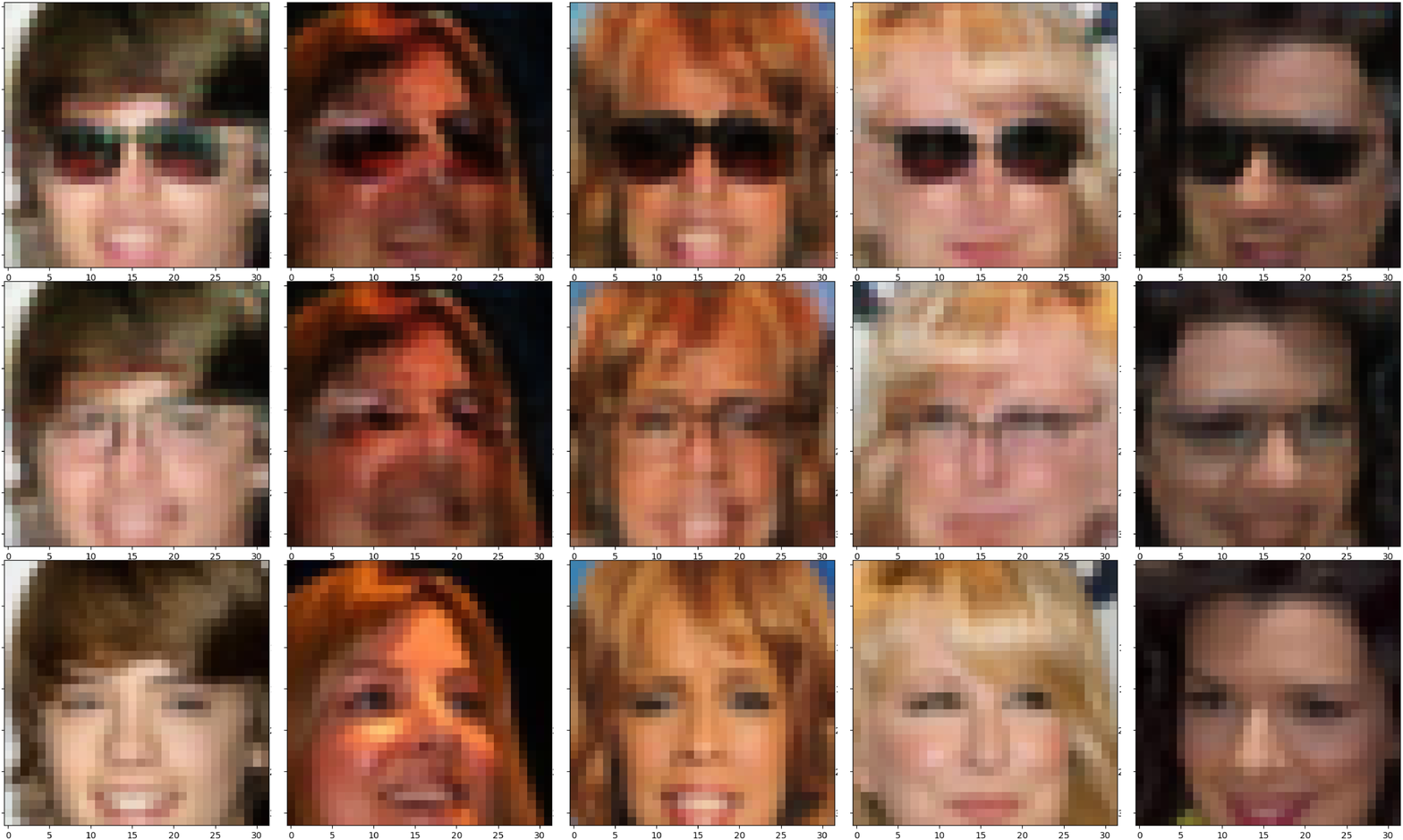}}
      \caption{Bottom row shows true samples $a_i \in D_E(A)$. Top two
        rows show the image $c(a_i, z_j)$ of adding glases with \textit{a specific latent
          vector} $z_1$ for the topmost row and $z_2$ for the middle row. Observe how the general style of the glasses stays the same in a given row, but gets adapted for every person that wears them.}
      \label{fig:z_fixed_montage}
\end{minipage}
\end{figure}

Just like in the case with standard GANs, we found training to be quite
unstable. Nevertheless, we did manage to train a model whose performance on our
tests of adding/removing glasses we now describe.
In Figure \ref{fig:add_change_glasses} (left) we show the model learns the task (i):
generating image of a specific person wearing glasses. Glasses are parameterized
by the latent vector $z \in D_E(B_Z)$. The model learns to warp the glasses and put them in the right angle and size, based on the shape of the face. This can especially be seen in Figure \ref{fig:z_fixed_montage}, where some of the faces are seen from an angle, but glasses still blend in naturally.
Figure \ref{fig:add_change_glasses} (right) shows the model learning task (ii):
\textit{changing} the glasses a person wears.

In Figure \ref{fig:remove_glasses} we see the model can learn to \textit{remove}
glasses. Observe how in some cases the model did not learn to remove the
glasses properly, as a slight outline of glasses can be seen. An interesting test of the learned semantics can be done by checking if a specific randomly sampled latent vector $z_j$ is consistent across different
images. Does the resulting image of the application of $g(a_i, z_j)$, contain
the same glasses as we vary the input image $a_i$?
The results for the tasks (ii, iii) are shown in Figure \ref{fig:z_fixed_montage}. It shows how the network has learned to associate a specific vector $z_j$ to a specific type of glasses and insert it in a natural way.

We note low diversity in generated glasses and a slight loss in image quality,
which is due to suboptimal architecture choice for neural networks.
Despite this, these experiments show that it is possible to train networks to (i) remove objects from, and (ii) parametrically insert objects into images in a \textit{unsupervised, unpaired fashion}.
Even though none of the networks were told that images contain people, glasses,
or objects of any kind, we highlight that they learned to preserve all the main facial
features.

\section{From categorical databases to deep learning}\label{sec:fdm_correspondence}

The formulation presented in this paper bears a striking and unexpected
similarity to Functorial Data Migration (FDM) \cite{FunctorialDataMigration}.
Given a \textit{categorical schema} $\qfr$ on some graph $G$, FDM defines a
functor category $\set^{\qfr}$ of database instance on that schema.
The notion of \textit{data integrity} is captured by \textit{path equivalence
  relations} which ensure any specified ``business rules'' hold.
The analogue of data integrity in neural networks is captured in the same way,
first introduced in CycleGAN \cite{CycleGAN} as \textit{cycle-consistency conditions}.
The main difference between the approaches is that in this paper we do not start
out with an implementation of the network instance functor, but rather we randomly
initialize it and then \textit{learn} it.

This shows that the underlying structures used for specifying data semantics for a given database systems are equivalent to the structures used to design data semantics which are possible to capture by training neural networks.

\section{Conclusion and future work}

In this paper we introduced a categorical formalism for training networks given
by an arbitrary categorical schema.
We showed there exists a correspondence between categorical formulation of databases and
neural network training. We developed a rudimentary theory of \textit{learning a
  specific class of functors using gradient descent.} 
Using the CelebA dataset we obtained experimental results and verified that
semantic image manipulation can be carried out in a novel way.

The category theory in this paper is only elementary and we believe there is
much more structure to be discovered. This work just scratching the surface of
the rich connection between machine learning and category theory. It opens up interesting avenues of research and it seems to be deserving of further exploration.

\bibliographystyle{eptcs}
\bibliography{references}

\appendix
\input{appendix}

\end{document}

%% file: preamble.tex
\usepackage[utf8]{inputenc}
\usepackage[english]{babel}
\usepackage[T1]{fontenc}

\usepackage{mathtools, amsmath, amssymb, xcolor,mathrsfs}
\usepackage{amsthm}
\usepackage{hyperref}
\usepackage{graphicx}
\usepackage{subcaption}
\usepackage{float}
\usepackage{faktor}
\usepackage{varwidth} 
\usepackage[percent]{overpic}

\usepackage{tikz}
\usepackage{lipsum}

\graphicspath{{./img/}}

\input{tikzstuff.tex}

\newcommand{\LMO}[2][over]{\ifthenelse{\equal{#1}{over}}{\overset{#2}{\bullet}}{\underset{#2}{\bullet}}}
\newcommand{\LTO}[2][\bullet]{\overset{\tn{#2}}{#1}}
\newcommand{\tn}[1]{\textnormal{#1}}
\newcommand{\boxCD}[2][black]{\fcolorbox{#1}{white}{\begin{varwidth}{\textwidth}\centering #2\end{varwidth}}}

\newcommand{\Cat}[1]{\mathbf{#1}}
\newcommand{\Fun}[1]{#1}

\newcommand{\RR}{\mathbb{R}}
\newcommand{\free}{\Cat{Free}}

\newcommand{\euc}{\mathbf{Euc}}
\newcommand{\loss}{\mathcal{L}}

\newcommand{\gen}[1]{\mathsf{Gen}_{#1}}

\newcommand{\set}{\Cat{Set}}
\newcommand{\zo}{[0, 1]^{100}}

\newcommand{\discr}[1]{\vert{#1}\vert}

\newcommand{\gimg}{[0, 1]^{64 \times 64 \times 3}}

\newcommand{\ih}{I_{H_p}}

\newcommand{\fr}{\free(G)}
\newcommand{\qfr}{\free(G)/{\sim}}
\newcommand{\vfr}{\discr{\fr}}

\newcommand{\para}{\Cat{Para}}

\newcommand{\ps}{\mathcal{P}}

\newcommand{\emb}{E}
\newcommand{\cpt}{\mathfrak{C}}

\newcommand{\bd}{\mathbf{D}}
\newcommand{\bg}{\mathbf{G}}

\newcommand{\arch}{\mathsf{Arch}}
\newcommand{\pspec}{\mathsf{PSpec}}
\newcommand{\model}{\mathsf{Model}}

\newtheorem{theorem}{Theorem}
\newtheorem{proposition}[theorem]{Proposition}
\newtheorem{example}[theorem]{Example}

\newtheorem{definition}[theorem]{Definition}

\makeatletter
\newcommand*{\doublerightarrow}[2]{\mathrel{
  \settowidth{\@tempdima}{$\scriptstyle#1$}
  \settowidth{\@tempdimb}{$\scriptstyle#2$}
  \ifdim\@tempdimb>\@tempdima \@tempdima=\@tempdimb\fi
  \mathop{\vcenter{
    \offinterlineskip\ialign{\hbox to\dimexpr\@tempdima+1em{##}\cr
    \rightarrowfill\cr\noalign{\kern.5ex}
    \rightarrowfill\cr}}}\limits^{\!#1}_{\!#2}}}
\newcommand*{\triplerightarrow}[1]{\mathrel{
  \settowidth{\@tempdima}{$\scriptstyle#1$}
  \mathop{\vcenter{
    \offinterlineskip\ialign{\hbox to\dimexpr\@tempdima+1em{##}\cr
    \rightarrowfill\cr\noalign{\kern.5ex}
    \rightarrowfill\cr\noalign{\kern.5ex}
    \rightarrowfill\cr}}}\limits^{\!#1}}}
\makeatother

%% file: tikzstuff.tex
\usepackage{tikz}

\usetikzlibrary{
	cd,
	shapes.geometric,
	decorations.markings,
	decorations.pathmorphing,
	positioning,
	arrows,
	shapes,
	calc,
	fit,
	quotes}

\tikzstyle{none}=[inner sep=0pt]
\tikzstyle{ibox}=[draw, rounded corners, minimum width = 30pt, minimum height =
18pt, thick]
\tikzstyle{update}=[->,>=stealth, very thick,decoration={snake, pre length =3pt, post
length =3pt},decorate]

\tikzset{
   oriented WD/.style={
      every to/.style={out=0,in=180,draw},
      label/.style={
         font=\everymath\expandafter{\the\everymath\scriptstyle},
         inner sep=0pt,
         node distance=2pt and -2pt},
      semithick,
      node distance=1 and 1,
      decoration={markings, mark=at position .5 with {\arrow{stealth};}},
      ar/.style={postaction={decorate}},
      execute at begin picture={\tikzset{
         x=\bbx, y=\bby,
         every fit/.style={inner xsep=\bbx, inner ysep=\bby}}}
      },
   bbx/.store in=\bbx,
   bbx = 1.5cm,
   bby/.store in=\bby,
   bby = 1.75ex,
   bb port sep/.store in=\bbportsep,
   bb port sep=2,
   bb port length/.store in=\bbportlen,
   bb port length=0pt,
   bb penetrate/.store in=\bbpenetrate,
   bb penetrate=0pt,
   bb min width/.store in=\bbminwidth,
   bb min width=1cm,
   bb rounded corners/.store in=\bbcorners,
   bb rounded corners=5pt,
   bb small/.style={bb port sep=1, bb port length=2.5pt, bbx=.4cm, bb min width=.4cm, bby=.7ex},
   bb/.code 2 args={
      \pgfmathsetlengthmacro{\bbheight}{\bbportsep * (max(#1,#2)) * \bby}
      \pgfkeysalso{draw,minimum height=\bbheight,minimum width=\bbminwidth,outer sep=0pt,
         rounded corners=\bbcorners,thick,
         prefix after command={\pgfextra{\let\fixname\tikzlastnode}},
         append after command={\pgfextra{\draw
            \ifnum #1=0{} \else foreach \i in {1,...,#1} {
               ($($(\fixname.north
	       west)+(0,.9\bbportsep)$)!{\i/(#1+1)}!($(\fixname.south
	       west)-(0,.9\bbportsep)$)$)
	       +(-\bbportlen,0) coordinate (\fixname_in\i) -- +(\bbpenetrate,0) coordinate (\fixname_in\i')}\fi 
            \ifnum #2=0{} \else foreach \i in {1,...,#2} {
               ($($(\fixname.north
	       east)+(0,\bbportsep)$)!{\i/(#2+1)}!($(\fixname.south
	       east)-(0,\bbportsep)$)$) +(-\bbpenetrate,0) coordinate (\fixname_out\i') -- +(\bbportlen,0) coordinate (\fixname_out\i)}\fi;
         }}}
   },
   bb name/.style={append after command={\pgfextra{\node[anchor=north] at
(\fixname.north) {#1};}}},
   ibb port sep/.store in=\ibbportsep,
   ibb port sep=2,
   ibb port length/.store in=\ibbportlen,
   ibb port length=4pt,
   ibb min width/.store in=\ibbminwidth,
   ibb min width=1cm,
   ibb rounded corners/.store in=\ibbcorners,
   ibb rounded corners=1pt,
   ibb small/.style={ibb port sep=1, ibb port length=2.5pt, bbx=.4cm, ibb min width=.4cm, bby=.7ex},
   ibb/.code 2 args={
	   \pgfmathsetlengthmacro{\ibbheight}{\ibbportsep * (max(#1,#2)) * \bby}
	   \pgfkeysalso{draw,color=gray!50,minimum height=\ibbheight,minimum width=\ibbminwidth,outer sep=0pt,
		   rounded corners=\ibbcorners,thick,
		   prefix after command={\pgfextra{\let\fixname\tikzlastnode}},
		   append after command={\pgfextra{\coordinate
			   \ifnum #1=0{} \else foreach \i in {1,...,#1} {
				   ($($(\fixname.north
					west)+(0,.9\ibbportsep)$)!{\i/(#1+1)}!($(\fixname.south
						west)-(0,.9\ibbportsep)$)$)
					   +(-\ibbportlen,0) coordinate (\fixname_in\i) -- +(\ibbportlen,0) coordinate (\fixname_in\i')}\fi 
					   \ifnum #2=0{} \else foreach \i in {1,...,#2} {
						   ($($(\fixname.north
							east)+(0,\ibbportsep)$)!{\i/(#2+1)}!($(\fixname.south
								east)-(0,\ibbportsep)$)$) +(-\ibbportlen,0) coordinate (\fixname_out\i') -- +(\ibbportlen,0) coordinate (\fixname_out\i)}\fi;
		   }}}
   },
   ibb name/.style={append after command={\pgfextra{\node[anchor=north] at
   (\fixname.north) {#1};}}},
   blankbb port sep/.store in=\blankbbportsep,
   blankbb port sep=2,
   blankbb min width/.store in=\blankbbminwidth,
   blankbb min width=1cm,
   blankbb rounded corners/.store in=\blankbbcorners,
   blankbb rounded corners=1pt,
   blankbb small/.style={blankbb port sep=1, blankbb port length=2.5pt, bbx=.4cm, blankbb min width=.4cm, bby=.7ex},
   blankbb/.code 2 args={
	   \pgfmathsetlengthmacro{\blankbbheight}{\blankbbportsep * (max(#1,#2)) * \bby}
	   \pgfkeysalso{draw,color=gray!50,minimum height=\blankbbheight,minimum width=\blankbbminwidth,outer sep=0pt,
		   rounded corners=\blankbbcorners,thick,
		   prefix after command={\pgfextra{\let\fixname\tikzlastnode}},
		   append after command={\pgfextra{\draw
			\ifnum #1=0{} \else foreach \i in {1,...,#1} {
			   ($($(\fixname.north
			   west)+(0,.9\ibbportsep)$)!{\i/(#1+1)}!($(\fixname.south
			   west)-(0,.9\ibbportsep)$)$)
					    coordinate (\fixname_in\i)}\fi 
			\ifnum #2=0{} \else foreach \i in {1,...,#2} {
			   ($($(\fixname.north
			   east)+(0,.9\ibbportsep)$)!{\i/(#2+1)}!($(\fixname.south
			   east)-(0,.9\ibbportsep)$)$) coordinate (\fixname_out\i)}\fi;
		   }}}
   },
   blankbb name/.style={append after command={\pgfextra{\node[anchor=north] at
     (\fixname.north) {#1};}}},
   symbb port sep/.store in=\symbbportsep,
   symbb port sep=2,
   symbb port length/.store in=\symbbportlen,
   symbb port length=0pt,
   symbb min width/.store in=\symbbminwidth,
   symbb min width=1cm,
   symbb rounded corners/.store in=\symbbcorners,
   symbb rounded corners=5pt,
   symbb small/.style={symbb port sep=1, symbb port length=2.5pt, symbbx=.4cm, symbb min width=.4cm, symbby=.7ex},
   symbb/.code 2 args={
      \pgfmathsetlengthmacro{\symbbheight}{\symbbportsep * (max(#1,#2)) * \bby}
      \pgfkeysalso{draw,minimum height=\symbbheight,minimum width=\symbbminwidth,outer sep=0pt,
         rounded corners=\symbbcorners,thick,
         prefix after command={\pgfextra{\let\fixname\tikzlastnode}},
         append after command={\pgfextra{\draw
            \ifnum #1=0{} \else foreach \i in {1,...,#1} {
               ($($(\fixname.north
	       west)+(0,.9\symbbportsep)$)!{\i/(#1+1)}!($(\fixname.south
	       west)-(0,.9\symbbportsep)$)$)
	       +(-\symbbportlen,0) coordinate (\fixname_in\i) -- +(\symbbportlen,0) coordinate (\fixname_in\i')}\fi 
            \ifnum #2=0{} \else foreach \i in {1,...,#2} {
               ($($(\fixname.north
	       east)+(0,.9\symbbportsep)$)!{\i/(#2+1)}!($(\fixname.south
	       east)-(0,.9\symbbportsep)$)$) +(-\symbbportlen,0) coordinate (\fixname_out\i') -- +(\symbbportlen,0) coordinate (\fixname_out\i)}\fi;
         }}}
   },
   symbb name/.style={append after command={\pgfextra{\node[anchor=north] at
(\fixname.north) {#1};}}},
}

%% file: appendix.tex
\section{Experiments}\label{ch:experiments}

In the experiments we have used optimizer Adam \cite{Adam} and the
Wasserstein GAN with gradient penalty \cite{ImprovedWGAN}. We used the
suggested choice of hyperparameters in \cite{ImprovedWGAN}.
The parameter $\gamma$ is set to $20$ and as such weighted the optimization procedure towards the path-equivalence, rather than the cycle-consistency loss.
All weights were initialized from a Gaussian distribution $\mathcal{N}(0, 0.01)$.
As suggested in \cite{ImprovedWGAN}, we always gave the discriminator a head start and trained it more, especially in the beginning. We set $n_{critic} = 50$ for the first 50 time steps and $n_{critic} = 5$ for all other time steps.

Discriminator $\bd(AB)$ and the discriminators for each $A$ and $B$ in $\bd(A
\times B)$ in first two experiments were $5$-layer ReLU convolutional neural networks
of type $\RR^q \times \RR^{32 \times 32 \times 3} \rightarrow \RR$. Kernel size
was set to $5$ and padding to $2$. We used stride $2$ to halve image size in all layers except the second, where we used stride $1$.
We used a fully-connected layer without any activations at the end of the
convolutional network to reduce the output size to $1$.


%% file: example.bbl
\begin{thebibliography}{10}
\providecommand{\bibitemdeclare}[2]{}
\providecommand{\surnamestart}{}
\providecommand{\surnameend}{}
\providecommand{\urlprefix}{Available at }
\providecommand{\url}[1]{\texttt{#1}}
\providecommand{\href}[2]{\texttt{#2}}
\providecommand{\urlalt}[2]{\href{#1}{#2}}
\providecommand{\doi}[1]{doi:\urlalt{http://dx.doi.org/#1}{#1}}
\providecommand{\bibinfo}[2]{#2}

\bibitemdeclare{article}{AugmentedCycleGAN}
\bibitem{AugmentedCycleGAN}
\bibinfo{author}{Amjad \surnamestart Almahairi\surnameend},
  \bibinfo{author}{Sai \surnamestart Rajeswar\surnameend},
  \bibinfo{author}{Alessandro \surnamestart Sordoni\surnameend},
  \bibinfo{author}{Philip \surnamestart Bachman\surnameend} \&
  \bibinfo{author}{Aaron~C. \surnamestart Courville\surnameend}
  (\bibinfo{year}{2018}): \emph{\bibinfo{title}{Augmented CycleGAN: Learning
  Many-to-Many Mappings from Unpaired Data}}.
\newblock {\sl \bibinfo{journal}{CoRR}} \bibinfo{volume}{abs/1802.10151}.
\newblock \urlprefix\url{http://arxiv.org/abs/1802.10151}.

\bibitemdeclare{article}{LTL}
\bibitem{LTL}
\bibinfo{author}{Marcin \surnamestart Andrychowicz\surnameend},
  \bibinfo{author}{Misha \surnamestart Denil\surnameend},
  \bibinfo{author}{Sergio~Gomez \surnamestart Colmenarejo\surnameend},
  \bibinfo{author}{Matthew~W. \surnamestart Hoffman\surnameend},
  \bibinfo{author}{David \surnamestart Pfau\surnameend}, \bibinfo{author}{Tom
  \surnamestart Schaul\surnameend} \& \bibinfo{author}{Nando \surnamestart
  de~Freitas\surnameend} (\bibinfo{year}{2016}): \emph{\bibinfo{title}{Learning
  to learn by gradient descent by gradient descent}}.
\newblock {\sl \bibinfo{journal}{CoRR}} \bibinfo{volume}{abs/1606.04474}.
\newblock \urlprefix\url{http://arxiv.org/abs/1606.04474}.

\bibitemdeclare{article}{WGAN}
\bibitem{WGAN}
\bibinfo{author}{Martin \surnamestart {Arjovsky}\surnameend},
  \bibinfo{author}{Soumith \surnamestart {Chintala}\surnameend} \&
  \bibinfo{author}{L{\'e}on \surnamestart {Bottou}\surnameend}
  (\bibinfo{year}{2017}): \emph{\bibinfo{title}{{Wasserstein GAN}}}.
\newblock {\sl \bibinfo{journal}{arXiv
  e-prints}}:\bibinfo{eid}{arXiv:1701.07875}.

\bibitemdeclare{inproceedings}{BackpropAsFunctor}
\bibitem{BackpropAsFunctor}
\bibinfo{author}{Brendan \surnamestart Fong\surnameend},
  \bibinfo{author}{David~I. \surnamestart Spivak\surnameend} \&
  \bibinfo{author}{R{\'{e}}my \surnamestart Tuy{\'{e}}ras\surnameend}
  (\bibinfo{year}{2019}): \emph{\bibinfo{title}{Backprop as Functor: {A}
  compositional perspective on supervised learning}}.
\newblock In: {\sl \bibinfo{booktitle}{34th Annual {ACM/IEEE} Symposium on
  Logic in Computer Science, {LICS} 2019, Vancouver, BC, Canada, June 24-27,
  2019}}, \bibinfo{publisher}{{IEEE}}, pp. \bibinfo{pages}{1--13},
  \doi{10.1109/LICS.2019.8785665}.

\bibitemdeclare{inproceedings}{CompositionalGameTheory}
\bibitem{CompositionalGameTheory}
\bibinfo{author}{Neil \surnamestart Ghani\surnameend}, \bibinfo{author}{Jules
  \surnamestart Hedges\surnameend}, \bibinfo{author}{Viktor \surnamestart
  Winschel\surnameend} \& \bibinfo{author}{Philipp \surnamestart
  Zahn\surnameend} (\bibinfo{year}{2018}): \emph{\bibinfo{title}{Compositional
  Game Theory}}.
\newblock In \bibinfo{editor}{Anuj \surnamestart Dawar\surnameend} \&
  \bibinfo{editor}{Erich \surnamestart Gr{\"{a}}del\surnameend}, editors: {\sl
  \bibinfo{booktitle}{Proceedings of the 33rd Annual {ACM/IEEE} Symposium on
  Logic in Computer Science, {LICS} 2018, Oxford, UK, July 09-12, 2018}},
  \bibinfo{publisher}{{ACM}}, pp. \bibinfo{pages}{472--481},
  \doi{10.1145/3209108.3209165}.

\bibitemdeclare{incollection}{GAN}
\bibitem{GAN}
\bibinfo{author}{Ian \surnamestart Goodfellow\surnameend},
  \bibinfo{author}{Jean \surnamestart Pouget-Abadie\surnameend},
  \bibinfo{author}{Mehdi \surnamestart Mirza\surnameend}, \bibinfo{author}{Bing
  \surnamestart Xu\surnameend}, \bibinfo{author}{David \surnamestart
  Warde-Farley\surnameend}, \bibinfo{author}{Sherjil \surnamestart
  Ozair\surnameend}, \bibinfo{author}{Aaron \surnamestart Courville\surnameend}
  \& \bibinfo{author}{Yoshua \surnamestart Bengio\surnameend}
  (\bibinfo{year}{2014}): \emph{\bibinfo{title}{Generative Adversarial Nets}}.
\newblock In \bibinfo{editor}{Z.~\surnamestart Ghahramani\surnameend},
  \bibinfo{editor}{M.~\surnamestart Welling\surnameend},
  \bibinfo{editor}{C.~\surnamestart Cortes\surnameend}, \bibinfo{editor}{N.~D.
  \surnamestart Lawrence\surnameend} \& \bibinfo{editor}{K.~Q. \surnamestart
  Weinberger\surnameend}, editors: {\sl \bibinfo{booktitle}{Advances in Neural
  Information Processing Systems 27}}, \bibinfo{publisher}{Curran Associates,
  Inc.}, pp. \bibinfo{pages}{2672--2680}.
\newblock
  \urlprefix\url{http://papers.nips.cc/paper/5423-generative-adversarial-nets.pdf}.

\bibitemdeclare{article}{ImprovedWGAN}
\bibitem{ImprovedWGAN}
\bibinfo{author}{Ishaan \surnamestart {Gulrajani}\surnameend},
  \bibinfo{author}{Faruk \surnamestart {Ahmed}\surnameend},
  \bibinfo{author}{Martin \surnamestart {Arjovsky}\surnameend},
  \bibinfo{author}{Vincent \surnamestart {Dumoulin}\surnameend} \&
  \bibinfo{author}{Aaron \surnamestart {Courville}\surnameend}
  (\bibinfo{year}{2017}): \emph{\bibinfo{title}{{Improved Training of
  Wasserstein GANs}}}.
\newblock {\sl \bibinfo{journal}{arXiv
  e-prints}}:\bibinfo{eid}{arXiv:1704.00028}.

\bibitemdeclare{article}{OnCompositionality}
\bibitem{OnCompositionality}
\bibinfo{author}{Jules \surnamestart Hedges\surnameend} (\bibinfo{year}{2017}):
  \emph{\bibinfo{title}{On Compositionality}}.
\newblock \urlprefix\url{https://julesh.com/2017/04/22/on-compositionality/}.

\bibitemdeclare{article}{SyntheticGradients}
\bibitem{SyntheticGradients}
\bibinfo{author}{Max \surnamestart Jaderberg\surnameend},
  \bibinfo{author}{Wojciech~Marian \surnamestart Czarnecki\surnameend},
  \bibinfo{author}{Simon \surnamestart Osindero\surnameend},
  \bibinfo{author}{Oriol \surnamestart Vinyals\surnameend},
  \bibinfo{author}{Alex \surnamestart Graves\surnameend} \&
  \bibinfo{author}{Koray \surnamestart Kavukcuoglu\surnameend}
  (\bibinfo{year}{2016}): \emph{\bibinfo{title}{Decoupled Neural Interfaces
  using Synthetic Gradients}}.
\newblock {\sl \bibinfo{journal}{CoRR}} \bibinfo{volume}{abs/1608.05343}.
\newblock \urlprefix\url{http://arxiv.org/abs/1608.05343}.

\bibitemdeclare{inproceedings}{Adam}
\bibitem{Adam}
\bibinfo{author}{Diederik~P. \surnamestart Kingma\surnameend} \&
  \bibinfo{author}{Jimmy \surnamestart Ba\surnameend} (\bibinfo{year}{2015}):
  \emph{\bibinfo{title}{Adam: {A} Method for Stochastic Optimization}}.
\newblock In: {\sl \bibinfo{booktitle}{{ICLR}}}.

\bibitemdeclare{inproceedings}{CelebA}
\bibitem{CelebA}
\bibinfo{author}{Ziwei \surnamestart Liu\surnameend}, \bibinfo{author}{Ping
  \surnamestart Luo\surnameend}, \bibinfo{author}{Xiaogang \surnamestart
  Wang\surnameend} \& \bibinfo{author}{Xiaoou \surnamestart Tang\surnameend}
  (\bibinfo{year}{2015}): \emph{\bibinfo{title}{Deep Learning Face Attributes
  in the Wild}}.
\newblock In: {\sl \bibinfo{booktitle}{2015 {IEEE} International Conference on
  Computer Vision, {ICCV} 2015, Santiago, Chile, December 7-13, 2015}},
  \bibinfo{publisher}{{IEEE} Computer Society}, pp.
  \bibinfo{pages}{3730--3738}, \doi{10.1109/ICCV.2015.425}.

\bibitemdeclare{book}{WorkingMathematician}
\bibitem{WorkingMathematician}
\bibinfo{author}{Saunders \surnamestart MacLane\surnameend}
  (\bibinfo{year}{1971}): \emph{\bibinfo{title}{Categories for the Working
  Mathematician}}.
\newblock \bibinfo{publisher}{Springer-Verlag}, \bibinfo{address}{New York}.
\newblock \bibinfo{note}{Graduate Texts in Mathematics, Vol. 5}.

\bibitemdeclare{article}{FunctorialDataMigration}
\bibitem{FunctorialDataMigration}
\bibinfo{author}{David~I. \surnamestart Spivak\surnameend}
  (\bibinfo{year}{2012}): \emph{\bibinfo{title}{Functorial data migration}}.
\newblock {\sl \bibinfo{journal}{Inf. Comput.}} \bibinfo{volume}{217}, pp.
  \bibinfo{pages}{31--51}, \doi{10.1016/j.ic.2012.05.001}.

\bibitemdeclare{article}{QueryLiftingProblems}
\bibitem{QueryLiftingProblems}
\bibinfo{author}{David~I. \surnamestart Spivak\surnameend}
  (\bibinfo{year}{2014}): \emph{\bibinfo{title}{Database queries and
  constraints via lifting problems}}.
\newblock {\sl \bibinfo{journal}{Math. Struct. Comput. Sci.}}
  \bibinfo{volume}{24}(\bibinfo{number}{6}), \doi{10.1017/S0960129513000479}.

\bibitemdeclare{article}{CycleGAN}
\bibitem{CycleGAN}
\bibinfo{author}{Jun{-}Yan \surnamestart Zhu\surnameend},
  \bibinfo{author}{Taesung \surnamestart Park\surnameend},
  \bibinfo{author}{Phillip \surnamestart Isola\surnameend} \&
  \bibinfo{author}{Alexei~A. \surnamestart Efros\surnameend}
  (\bibinfo{year}{2017}): \emph{\bibinfo{title}{Unpaired Image-to-Image
  Translation using Cycle-Consistent Adversarial Networks}}.
\newblock {\sl \bibinfo{journal}{CoRR}} \bibinfo{volume}{abs/1703.10593}.
\newblock \urlprefix\url{http://arxiv.org/abs/1703.10593}.

\end{thebibliography}
